\documentclass{Interspeech2024}
\usepackage{mathtools}
\usepackage{multirow}
\usepackage{footmisc}

\usepackage{hyperref}
\usepackage{url}

\interspeechcameraready

\title{LI-TTA: Language Informed Test-Time Adaptation for Automatic Speech Recognition}

\name[affiliation={1*}]{Eunseop}{Yoon}
\name[affiliation={1*}]{Hee Suk}{Yoon}
\name[affiliation={2}]{John}{Harvill}
\name[affiliation={2}]{Mark}{Hasegawa-Johnson}
\name[affiliation={1**}]{Chang}{D. Yoo}

\address{
  $^1$Korea Advanced Institute of Science and Technology (KAIST), Republic of Korea\\
  $^2$University of Illinois at Urbana-Champaign, USA}
\email{\{esyoon97, hskyoon, cd\_yoo\}@kaist.ac.kr, 
        \{harvill2, jhasegaw\}@illinois.edu}


\begin{document}
\maketitle

\begin{abstract}
Test-Time Adaptation (TTA) has emerged as a crucial solution to the domain shift challenge, wherein the target environment diverges from the original training environment. A prime exemplification is TTA for Automatic Speech Recognition (ASR), which enhances model performance by leveraging output prediction entropy minimization as a self-supervision signal. However, a key limitation of this self-supervision lies in its primary focus on acoustic features, with minimal attention to the linguistic properties of the input. To address this gap, we propose Language Informed Test-Time Adaptation (LI-TTA), which incorporates linguistic insights during TTA for ASR. LI-TTA integrates corrections from an external language model to merge linguistic with acoustic information by minimizing the CTC loss from the correction alongside the standard TTA loss. With extensive experiments, we show that LI-TTA effectively improves the performance of TTA for ASR in various distribution shift situations. The code is publicly accessible at \href{https://github.com/EsYoon7/LiTTA}{https://github.com/EsYoon7/LiTTA}. \let\thefootnote\relax\footnotetext{* Equal Contribution}\let\thefootnote\relax\footnotetext{** Corresponding Author}
\end{abstract}

\section{Introduction}

Automatic Speech Recognition (ASR) systems have seen remarkable advancements through the development of pre-trained models. These pre-trained models have significantly improved ASR performance by leveraging vast amounts of data to capture complex patterns in speech. However, in real-world applications, ASR models frequently encounter inputs with distributions that differ from their training datasets, leading to performance degradation. This issue is exacerbated by the practical impossibility and computational expense of continuously collecting and training on new data for every possible distribution shift encountered in real-world scenarios. 

To address this challenge, Test-Time Adaptation (TTA) has emerged as a promising solution, drawing considerable attention for its ability to adapt models at inference time without the need for additional training data. Drawing inspiration from Test-Time Adaptation (TTA) strategies employed in computer vision \cite{tta_tent,tta_related,tta_related2_memo,tta_related3_tpt,tta_related4_ctpt}, recent methods \cite{suta, sgem} have demonstrated noteworthy effectiveness in adapting pre-trained ASR models \cite{wav2vec2, conformer, transducer} by utilizing entropy minimization as a self-supervision signal. 

\begin{figure}[t]
	\centering
    	\includegraphics[width=0.9\linewidth]{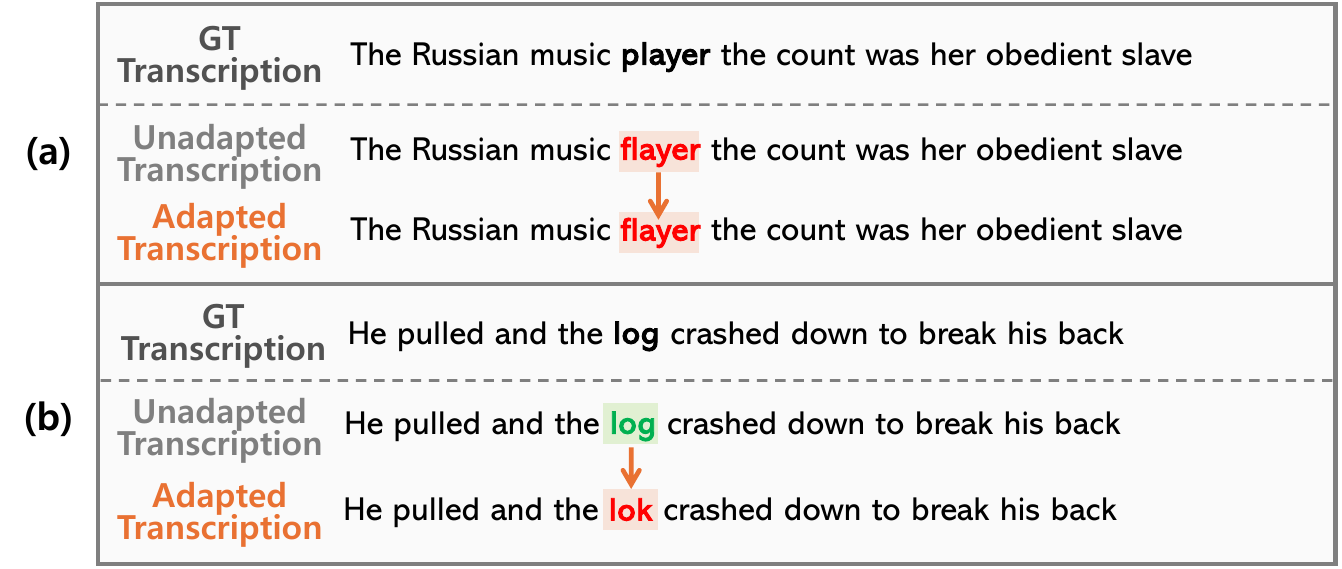}
    \caption{\textbf{Examples of failure cases with traditional TTA methods in ASR due to the absence of linguistic feedback.} (a) Predictions with phonetically similar words persist incorrectly, misaligning with the sentence context. (b) Additionally, some accurate predictions are incorrectly replaced with contextually unfit words of similar phonetic structure.}

	\label{fig:intro}
\end{figure}

A significant limitation of the self-supervision signals in ASR models is their excessive focus on acoustic features, frequently neglecting the linguistic aspects of spoken language. While external language models are commonly employed during the beam search decoding phase to process logits from ASR models, no feedback from these language models can backpropagate through the ASR system to aid the TTA process. For example, Figure \ref{fig:intro} illustrates failure cases of traditional TTA methods where the absence of linguistic feedback leads to two main issues: phonetically similar but contextually inappropriate predictions remain uncorrected (Figure \ref{fig:intro}-(a)), and in some instances, initially correct predictions are erroneously altered to phonetically similar words that do not suit the context of the sentence (Figure \ref{fig:intro}-(b)). These examples underscore the necessity for integrating linguistic information into the TTA process to enhance both the accuracy and contextual relevance of ASR predictions.

In response to this challenge, this paper presents Language Informed Test-Time Adaptation (LI-TTA) which integrates linguistic guidance into the TTA process for ASR systems. By employing corrections derived from an external instructed-tuned language model, LI-TTA effectively combines linguistic and acoustic information. This integration is achieved through the joint minimization of the Connectionist Temporal Classification (CTC) \cite{ctc} loss, resulting from these corrections, and the conventional TTA loss. Our comprehensive evaluations across several domain shift benchmarks demonstrate the efficacy of LI-TTA: it not only enhances the precision of ASR-generated transcriptions, as evidenced by lower word error rates but also ensures their contextual fidelity, as indicated by reduced perplexity scores.

\section{Related Work}

\subsection{Distribution shift on Automatic Speech Recognition}

Recent advancements in Automatic Speech Recognition (ASR) models have achieved surprising performance on in-distribution training data, primarily attributed to the methodology of self-supervised pre-training followed by supervised fine-tuning \cite{wav2vec, wav2vec2, conformer, hubert}. During the pre-training phase, models comprehensively understand acoustic features from audio inputs without necessitating labeled data. This foundational knowledge facilitates the subsequent fine-tuning stage, wherein models are explicitly trained to correlate audio inputs with their corresponding transcription texts. Despite these advancements, the inherent variability of speech recognition—attributable to diverse contexts, speaker characteristics, and background noises—presents significant challenges. 
Models frequently encounter samples from distributions unseen during training, leading to degraded performance in real-world applications.

Addressing the issue of distribution shift, \cite{da_asr1_augmenation, da_asr2_alignment, asr_accent_training1, asr_accent_training2} proposed a strategy of fine-tuning ASR models using target domain training data. Although this approach demonstrates effectiveness in aligning the model with the target domain, it is computationally inefficient. In response, \cite{intapt} introduced a prompt tuning approach to adapt the model to the target domain, which not only enhances model performance in new domains but also preserves efficacy in the source domain. A commonality among these strategies is their reliance on domain-specific training data and the necessity of a training process to achieve improvements in the target domain.

In light of these limitations, the focus has recently shifted towards test-time adaptation to optimize ASR models for target domains without the need for extensive training data or following training procedures. This approach necessitates only minimal adaptation steps during inference to achieve significant improvements \cite{suta, sgem}.

\subsection{Test-time adaptation}

Test-Time Adaptation (TTA) enhances deep learning model performance in new domains by adapting the model during inference without requiring labeled training data. In the field of computer vision, strategies such as entropy minimization \cite{tta_tent}, batch normalization \cite{ttn}, and data augmentation \cite{tta_related2_memo} have been proposed, demonstrating effectiveness in image classification. Recent research has also extended TTA to vision-language models, highlighting the approach's versatility and potential across different domains \cite{tta_related3_tpt, tta_related4_ctpt}. Despite the growing attention of TTA research in computer vision, its application within the speech domain remains comparatively sparse.

To address this gap, \cite{suta} pioneered the application of TTA to Automatic Speech Recognition (ASR), utilizing a single-utterance adaptation method. Further enhancing this approach, \cite{sgem} introduced a beam search strategy to improve performance during adaptation. Nevertheless, the principal drawback of these methods is their overwhelming emphasis on acoustic aspects, largely overlooking the linguistic dimensions of the input. In response to this limitation, we introduce Language Informed Test-Time Adaptation (LI-TTA), a method that integrates linguistic attributes into the TTA framework for ASR. 

\begin{figure}[t]
	\centering
    	\includegraphics[width=1.0\linewidth]{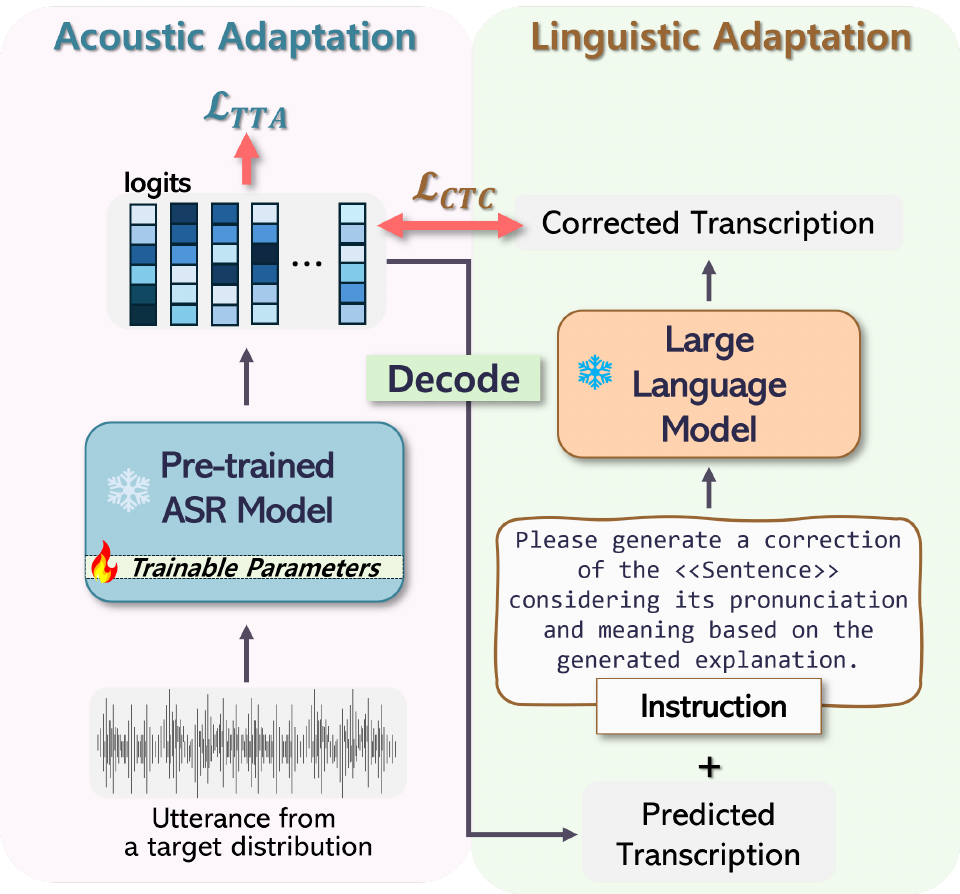}
	\caption{\textbf{An overview of our proposed Language Informed Test-Time Adaptation (LI-TTA)}. LI-TTA integrates corrections from an external instruction-tuned language model to merge linguistic with acoustic information by minimizing the CTC loss from the correction alongside the standard TTA loss.}
	\label{fig:method}
\end{figure}
\section{Method}
In this section, we present Language Informed Test-Time Adaptation (LI-TTA), a novel approach that enhances Test-Time Adaptation (TTA) for Automatic Speech Recognition (ASR) by integrating linguistic insights from an external language model. We begin with Section \ref{method:preliminary}, where the setup for TTA in ASR contexts is outlined. In Section \ref{method:ppl}, we show that the existing TTA framework fails to uphold linguistic knowledge, although using a language model during the decoding phase. Following this, in Section \ref{method:litta}, we provide a comprehensive explanation of our proposed LI-TTA. 
\subsection{Preliminary}
\label{method:preliminary}
\noindent\textbf{General Automatic Speech Recognition (ASR)}\quad Consider an ASR model, denoted by \(f(\cdot | \theta)\), which is trained using labeled data from a source domain, represented as \(\mathcal{D}_{\text{source}} = \{(x_i^{\text{source}}, \mathbf{y}_i^{\text{source}})\}_{i=1}^N\). This model processes a raw audio waveform \(x\), producing logits \(f(x | \theta) \in \mathbb{R}^{L \times C}\) across \(L\) timesteps for a vocabulary \(\mathcal{V}\) of size \(C\), where \(\theta\) denotes the model’s parameters. The model aims to calculate the joint log probability, \(\log p(\mathbf{y} | x, \theta)\), for a potential transcription \(\mathbf{y} = (y_{(1)}, \ldots, y_{(L)})\) using an autoregressive approach:
\begin{align}
& \log p(\mathbf{y} | x, \theta) = \log p_{\text{AM}} (\mathbf{y}| x, \theta) + \lambda \log p_{\text{LM}}(\mathbf{y}) \\
&= \sum_{t=1}^{L} [\log p_{\text{AM}} (y_{(t)} | y_{(<t)}, x, \theta) + \lambda \log p_{\text{LM}}(y_{(t)} | y_{(<t)})], \nonumber
\end{align}

where $y_{(t)} \in \mathcal{V}$. Here, \(p_{\text{AM}}(\mathbf{y}| x, \theta)\) refers to the probability as predicted by the ASR model for the input \(x\), whereas \(p_{\text{LM}}(\mathbf{y})\) represents the likelihood according to an autoregressive language model (LM). 
The parameter \(\lambda\) serves to balance the influence of the LM. \textit{It is important to note that this LM is utilized solely during the decoding stage of inference to guide the selection of the optimal solution, denoted as $\mathbf{y^*}$.}

\hfill \break
\noindent\textbf{Test-Time Adaptation (TTA) for ASR}\quad In the context of TTA for ASR, the objective is to tailor the model \(f(\cdot | \theta)\) to operate optimally within the domain of unlabeled target speech, denoted by \(\mathcal{D}_\text{target} = \{x_i^\text{target}\}_{i=1}^M\). This adjustment process particularly focuses on a scenario termed as the \textit{single-utterance TTA} approach \cite{suta, sgem}, where individual utterance \(x_i^\text{target}\) from \(\mathcal{D}_\text{target}\) is used to fine-tune \(f(\cdot | \theta)\). The aim is to refine the model's predictions, achieving enhanced prediction logits \(\log p(\mathbf{y}|x_i^\text{target}, \theta)\) by entropy minimization. Since \(p_{\text{LM}}\) is independent of \(\theta\), the adaptation process is influenced solely by \(p_{\text{AM}}\), highlighting the model's reliance on acoustic modeling adjustments during TTA.
\begin{figure}[t]
	\centering
    	\includegraphics[width=0.84\linewidth]{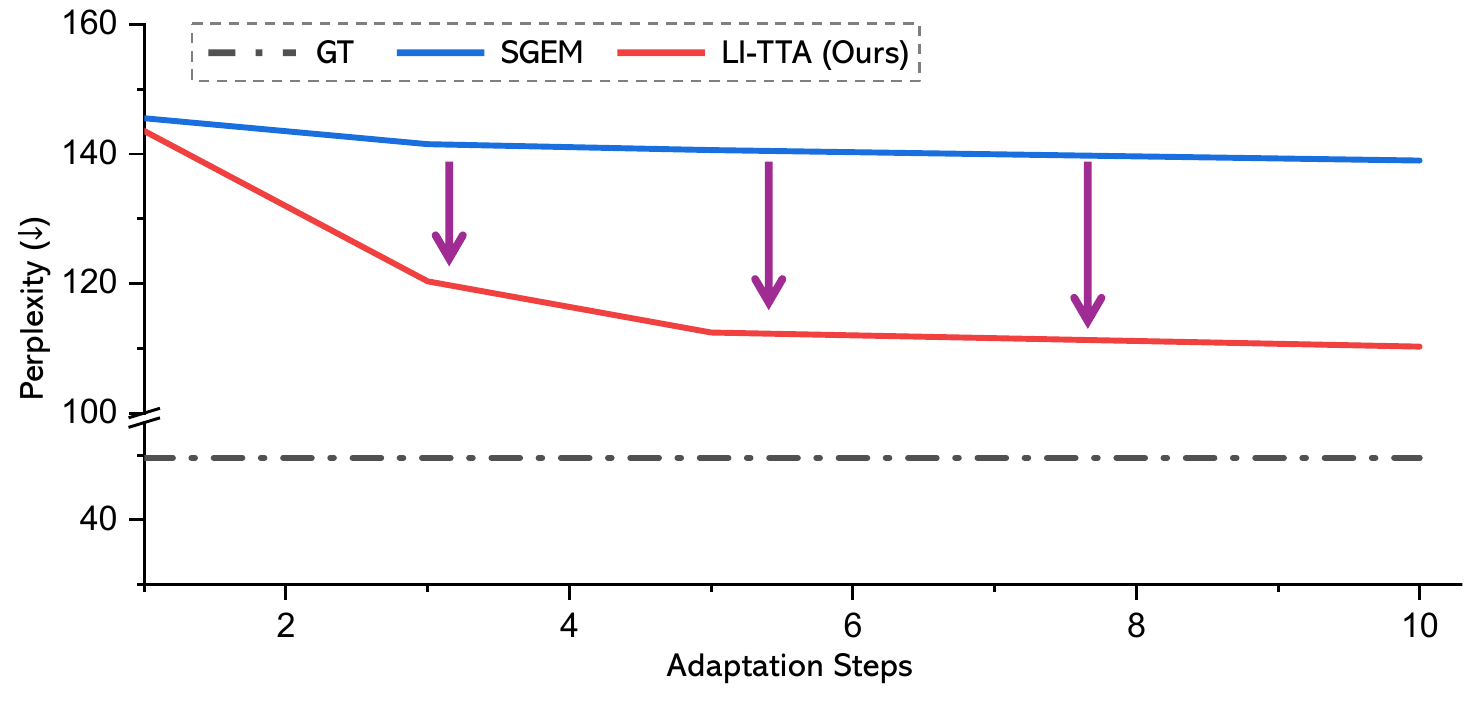}
	\caption{\textbf{The trend of perplexity (PPL) with respect to the TTA adaptation steps}. Traditional TTA approach (SGEM \cite{sgem}) shows only marginal reductions in perplexity due to the absence of linguistic feedback during the adaptation process. Conversely, our proposed LI-TTA demonstrates a significant decrease in perplexity, benefiting from the integration of linguistic insights.}
 \vspace{-20pt}
	\label{fig:ppl}
\end{figure}
\subsection{Lack of Linguistic Supervision of Existing TTA}
\label{method:ppl}
As discussed in Section \ref{method:preliminary}, \textit{while external language models play a crucial role during the beam search decoding phase of ASR models by processing logits, they are unable to provide feedback that can backpropagate through the ASR system to enhance the TTA process.} This constraint is emphasized in Figure \ref{fig:ppl}, where sentence semantic coherence and grammatical accuracy are evaluated using perplexity (PPL) scores from an advanced external LM (GPT-2 \cite{gpt}). Lower PPL scores reflect the greater confidence of LM in the grammatical and semantic integrity of the sentence, whereas higher scores indicate potential issues in terms of the coherence and precision of the sentence. Our observations reveal that increasing the number of adaptation steps does not necessarily lead to a reduction in PPL scores for the previous TTA method (SGEM \cite{sgem}). This leads us to propose LI-TTA in the subsequent section, aiming to provide more effective linguistic feedback during TTA, thereby achieving PPL scores that closely align with the ground truth (GT).

\subsection{LI-TTA: Language Informed TTA for ASR}
\label{method:litta}
This section presents LI-TTA, depicted in Figure \ref{fig:method}, which leverages an external instruction-tuned language model to refine ASR model predictions within TTA framework through CTC loss minimization. In this process, an ASR model represented as $f(\cdot | \theta)$, generates an initial prediction $\mathbf{y}$ and decoded transcription $\mathbf{y'}$. Subsequently, this transcription is fed to an external instruction-tuned LM ($\text{LM}_{\text{external}}$)\footnote{Note that this language model is different from the one used during the beam search decoding phase explained in Section \ref{method:preliminary}.}, which, based on a specific set of instructions, produces a corrected version $\tilde{\mathbf{y}}$:
\begin{align}
\tilde{\mathbf{y}} = \text{LM}_{\text{external}}(\text{Instruction}, \mathbf{y'}).
\end{align}

Given that the original prediction $\mathbf{y}$ and the corrected transcription $\tilde{\mathbf{y}}$ might not align perfectly, we employ the CTC loss \cite{ctc} as a means of linguistic feedback to enhance the ASR model. The finetuning of $f(\cdot | \theta)$ is thus guided by a composite loss function:
\begin{align}
\mathcal{L} = \mathcal{L}_\text{TTA} + \lambda_{\text{LI}} \cdot \mathcal{L}_{\text{CTC}}(\tilde{\mathbf{y}}, \mathbf{y}),
\end{align}
where $\mathcal{L}_\text{TTA}$ represents the original TTA loss \cite{sgem} and $\lambda_{\text{LI}}$ is a weighting factor that adjusts the influence of the linguistic feedback on the overall loss. Such an approach allows linguistic information to guide the TTA process, thereby allowing the predicted transcription to have a better coherent sentence structure.

\section{Experimental Setting}
\noindent\textbf{Model}\quad For the evaluation of LI-TTA we use the wav2vec 2.0 model\footnote{\scriptsize{https://huggingface.co/facebook/wav2vec2-base-960h}} \cite{wav2vec2}, which was trained on the LibriSpeech dataset \cite{libri}. For the language model used during the beam search decoding phase, we utilize the external 4-gram language model\footnote{\scriptsize{https://huggingface.co/patrickvonplaten/wav2vec2-base-100h-with-lm}}. For the external instruction-tuned language model ($\text{LM}_{\text{external}}$), we use OpenChat 3.5\footnote{\scriptsize{https://huggingface.co/openchat/openchat\_3.5}}.

\hfill \break
\noindent\textbf{Datasets}\quad To assess LI-TTA's adaptability across different domain shifts, we conducted tests on a variety of datasets. For evaluating performance against unseen speakers and vocabulary, we use the test set from TED-LIUM 2 (TD) \cite{dataset_TED}, Common Voice (CV), and \cite{dataset_cv}. Furthermore, to examine LI-TTA's robustness against background noise, we introduced eight noise conditions (air conditioner (AC), airport announcement (AA), babble (BA), copy machine (CM), munching (MU), neighbors (NB), shutting door (SD), and typing (TP)) at an SNR of 10dB into the LibriSpeech test-other dataset \cite{libri}. These noise samples were drawn from the MS-SNSD noise dataset \cite{dataset_mssnsd}. Lastly, the L2-Artic corpus \cite{dataset_l2artic}, consisting of non-native English speakers, served to test LI-TTA's efficacy in handling pronounced accent and pronunciation variations, with a focus on a single speaker from each first language category. To underscore the effectiveness of LI-TTA in ensuring that transcriptions accurately reflect the spoken content, our evaluation prioritizes test sets where the ground truth (GT) perplexity, as measured by GPT-2, falls below 70. This criterion helps to selectively focus on those datasets that present meaningful sentence contexts, efficiently filtering out test sets characterized by a lack of coherent sentence structure.
\begin{table*}[!t]
\scriptsize
\centering
\caption{Comparison of TTA performance, measured using word error rate (WER) ($\downarrow$) and perplexity (PPL) ($\downarrow$), across 11 datasets with various types of domain shifts. The results are obtained using greedy decoding for inference. Bold represents the best scores.}
\label{table:main}
\vspace{-0.1in}
\begin{tabular}{lcccccccccccccc}
\toprule
     \textbf{Method} & &\textbf{TD} & \textbf{CV}  & \textbf{AC} & \textbf{AA} & \textbf{BA} & \textbf{CM} & \textbf{MU} & 
     \textbf{NB} & \textbf{SD} & \textbf{TP} & \textbf{VC} & \textbf{Avg.} \\ 
     \midrule
      \multirow{2}{*}{Unadapted} & WER & 11.72 & 26.08  & 21.17 & 32.45 & 60.05 & 42.38 & 45.33 & 108.81 & 14.07 & 18.30 & 51.26 & 39.24 \\
                                 & PPL & 485.66 & 689.39  & 413.56 & 554.97 & 1101.00 & 1086.52 & 942.13 & 1333.51 & 212.45 & 290.10 & 1490.31 & 781.78 \\ 
      \midrule
     \multirow{2}{*}{SUTA \cite{suta}} & WER & 10.53  & 21.97 & 13.85 & 31.53 & 49.82 & 33.29 & 34.61 & 102.59 & 10.66 & 12.01 & 36.49 & 32.49 \\ 
                            & PPL & 358.67  & 447.86 & 545.96 & 512.23  & 1029.24 & 625.73 & 1358.04 & 1008.84 & 169.20  & 127.34 & 1138.00 & 665.56 \\ 
   \midrule
     \multirow{2}{*}{SGEM \cite{sgem}} & WER & 10.33  & 21.31 & 12.53 & 32.45 & 48.20 & 32.44 & 33.76 & 101.0 & 10.46 & 11.72 & 34.11 & 31.66 \\ 
                            & PPL & 297.37  & 471.61 & 630.62 & 507.67 & 1034.70 & 564.48 & 1468.38 & 854.44 & 128.27 & 135.23 & 1284.91 & 670.70 \\ 
    \midrule
     \multirow{2}{*}{LI-TTA (ours)} & WER & \textbf{10.30} & \textbf{20.35} & \textbf{12.16} & \textbf{24.55} & \textbf{46.55} & \textbf{31.88} & \textbf{32.31} & \textbf{99.58} & \textbf{10.44} & \textbf{11.33} & \textbf{33.23} & \textbf{30.24} \\
                & PPL & \textbf{271.18} & \textbf{389.76} & \textbf{412.74} & \textbf{444.65} & \textbf{803.42} & \textbf{563.37} & \textbf{1262.94} & \textbf{812.31} & \textbf{122.71} & \textbf{120.49} & \textbf{1098.82} & \textbf{572.94} \\     
\bottomrule
\vspace{-0.2in}
\end{tabular}
\end{table*}

\hfill \break
\noindent\textbf{Implementation}\quad For the hyperparameters we follow \cite{sgem} which optimized hyperparameters on CHiME-3 \cite{dataset_chime}. Specifically, we use the AdamW optimizer \cite{adamw}, coupled with a cosine annealing learning rate scheduler, setting the initial and final learning rates to $4 \cdot 10^{-5}$ and $2 \cdot 10^{-5}$, respectively. The number of TTA adaptation steps is set to 10, applying SGEM \cite{sgem} to compute $\mathcal{L}_{\text{TTA}}$. The decoding stage hyperparameter $\lambda$ is set to 0.3. For $\lambda_{\text{LI}}$, we use an adaptive strategy, setting it as $\frac{\mathcal{L}_{\text{TTA}}}{\mathcal{L}_{\text{TTA}} + \mathcal{L}_{\text{CTC}}}$. The external language model ($\text{LM}_{\text{external}}$) receives instructions formulated as \texttt{"Please generate a correction of the <<sentence>> considering the pronunciation and overall context"}. We only train feature extractors of the ASR model. All experiments are conducted on NVIDIA A100 80GB PCIe.

\begin{table}[h!]
\scriptsize
\centering
\caption{Comparison of TTA performance, measured using word error rate (WER) ($\downarrow$) and perplexity (PPL) ($\downarrow$) on six non-native English speech corpora.}
\label{table:non_native}
\begin{tabular}{lccccc}
\toprule
    \textbf{Setting} & & Unadapted & SUTA \cite{suta} & SGEM \cite{sgem} & LI-TTA \\ 
    \midrule
    \multirow{2}{*}{Arabic} & WER & 13.98 & 12.11 & 11.93 & \textbf{10.42} \\
     & PPL & 412.75 & 363.78 & 358.28 & \textbf{317.56} \\
    \midrule
    \multirow{2}{*}{Mandarin} & WER & 14.28 & 12.81 & 12.42 & \textbf{10.69} \\
     & PPL & 219.38 & 168.14 & 161.59 & \textbf{141.78} \\
    \midrule
    \multirow{2}{*}{Hindi} & WER & 11.54 & 10.40 & 9.86 & \textbf{9.08} \\
     & PPL & 190.16 & 182.38 & 177.24 & \textbf{146.52} \\
    \midrule
    \multirow{2}{*}{Korean} & WER & 10.65 & 8.96 & 8.21 & \textbf{7.86} \\
    & PPL & 210.96 & 159.68 & 182.00 & \textbf{143.69} \\
    \midrule
    \multirow{2}{*}{Spanish} & WER & 19.02 & 12.68 & 13.02 & \textbf{11.73} \\
     & PPL & 529.71 & 202.24 & 252.68 & \textbf{159.14} \\
    \midrule
    \multirow{2}{*}{Vietnamese}& WER & 34.92 & 30.83 & 31.06 & \textbf{28.80} \\ 
    & PPL & 988.28 & 733.11 & 651.78 & \textbf{591.65} \\
    \midrule
    \midrule
    \multirow{2}{*}{\textbf{Average}}& WER  & 17.40 & 14.63 & 14.41 &  \textbf{13.10} \\
     & PPL & 425.21 & 301.55 & 297.26 & \textbf{250.06} \\
\bottomrule
\vspace{-0.25in}
\end{tabular}
\end{table}
\section{Result}

\subsection{Main Results}
We compare the TTA performance of the ASR model across 11 datasets with various domain shifts. The results, detailed in Table \ref{table:main}, include both the word error rate (WER) and perplexity (PPL) metrics for the ASR model's outputs, with PPL scores derived using the GPT-2 \cite{gpt} model. Notably, the incorporation of LI-TTA into the ASR models consistently yields the lowest word error rate (WER) of target utterances with an average of 30.24\%. Furthermore, LI-TTA also achieves the lowest perplexity (PPL) across all 11 datasets, with an average of 572.94. This indicates the efficacy of our unsupervised objectives, which significantly refine the accuracy of ASR-generated transcriptions—as evidenced by reduced WER—and ensure their contextual fidelity, as demonstrated by lower PPL scores.

\subsection{Non-Native English Speech Corpora}
To further demonstrate the effectiveness of LI-TTA across domain shifts, we conducted evaluations on six distinct corpora comprising non-native English speech. The results in Table \ref{table:non_native} show that LI-TTA consistently surpasses the baselines across all tested corpora in terms of both WER and PPL, highlighting its robustness to significant pronunciation and accent variations.
\subsection{Qualitative Results}
\begin{figure}[t]
	\centering
    	\includegraphics[width=\linewidth]{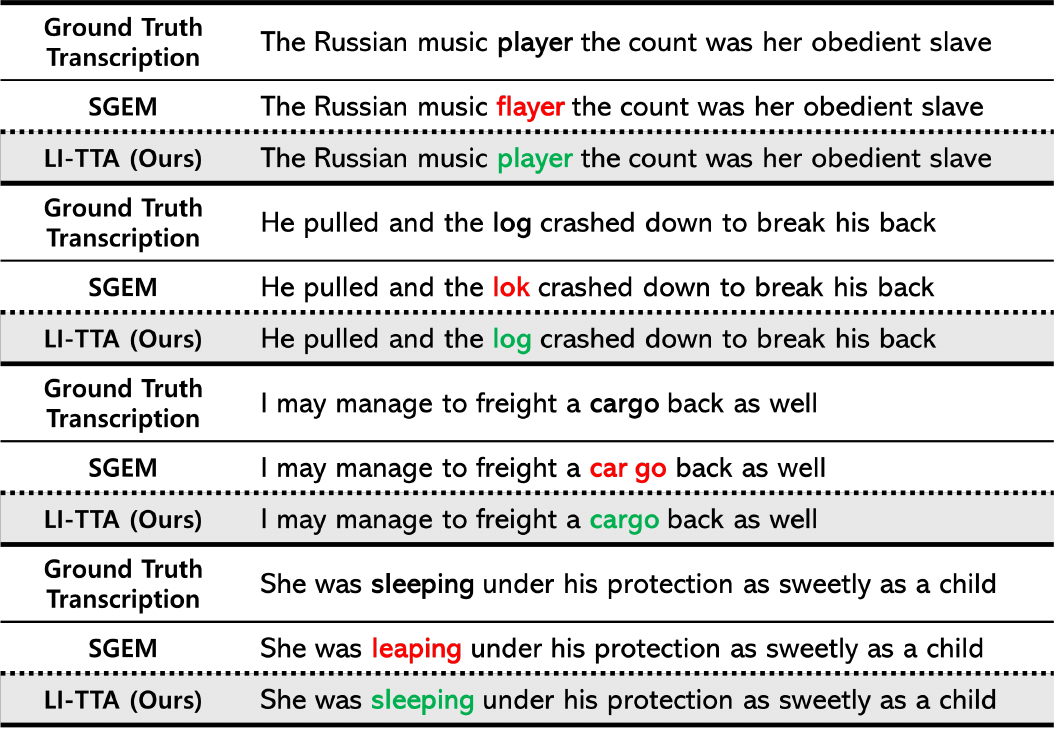}
	\caption{\textbf{Examples of failure cases of previous TTA method (SGEM \cite{sgem}) and the prediction from our proposed LI-TTA with contextually correct transcriptions.}} 
	\label{fig:qual}
 \vspace{-2pt}
\end{figure}

In Figure \ref{fig:qual}, we present a qualitative comparison between the conventional TTA method, SGEM \cite{sgem}, and our proposed LI-TTA through specific examples that highlight their prediction differences. These examples illustrate instances where SGEM fails to correctly interpret the context of the utterance, leading to erroneous transcriptions. Conversely, LI-TTA demonstrates its adeptness at navigating such complexities, accurately capturing the intended context, and producing contextually appropriate transcriptions. This distinction highlights the advantage of incorporating linguistic knowledge into the TTA process and underscores the inadequacy of methods like SGEM that depend exclusively on acoustic features. 

\section{Conclusions}
In this study, we tackled the domain shift challenge in ASR by proposing Language Informed Test-Time Adaptation (LI-TTA), a novel method that enriches Test-Time Adaptation (TTA) with linguistic insights. By integrating corrections from an external language model, LI-TTA not only mitigates the limitations of traditional TTA approaches that focus predominantly on acoustic features but also significantly enhances ASR accuracy and contextual correctness. Our comprehensive evaluations across diverse domain shifts show the effectiveness of LI-TTA, evidenced by improved word error rates and perplexity scores. 

\section{Limitation}
 In this paper, we show the effectiveness of LI-TTA on a CTC-based ASR model. Future work can extend the validation of LI-TTA's efficacy across diverse ASR architectures, including conformer-based and transducer-based models, to ascertain its adaptability in a broader array of speech recognition systems.

\section{Acknowledgements}

This work was partially supported by Institute for Information \& communications Technology Planning \& Evaluation (IITP) grant funded by the Korea government(MSIT) (No. 2021-0-01381, Development of Causal AI through Video Understanding and Reinforcement Learning, and Its Applications to Real Environments) and SAMSUNG Research, Samsung Electronics Co.,Ltd.

\newpage
\bibliographystyle{IEEEtran}
\bibliography{mybib}

\end{document}